\pdfoutput=1

\documentclass[11pt]{article}

\usepackage{acl}

\usepackage{times}
\usepackage{latexsym}

\usepackage[T1]{fontenc}

\usepackage[utf8]{inputenc}

\usepackage{microtype}
\usepackage{amsmath}
\usepackage{graphicx}
\usepackage{amsfonts}
\usepackage{enumitem}

\newcommand{\longdupe}{Deduction under Perturbed Evidence }
\newcommand{\longdupens}{Deduction under Perturbed Evidence}
\newcommand{\dupe}{DUPE }
\newcommand{\dupens}{DUPE}
\newcommand{\duped}{DUPEd }
\newcommand{\dupedns}{DUPEd}
\newcommand{\DUPED}{DUPEd }
\newcommand{\DUPEDns}{DUPEd}
\newcommand{\GPT}{LLMs }
\newcommand{\GPTns}{LLMs}

\newcommand\tr{\textrm}

\title{\longdupens:\\
Probing Student Simulation Capabilities of Large Language Models
}

\author{Shashank Sonkar \\
  Rice University \\
  \texttt{ss164@rice.edu} \\\And
  Richard G. Baraniuk \\
  Rice University \\
  \texttt{richb@rice.edu}
  }

\begin{document}
\maketitle

\begin{abstract}
We explore whether Large Language Models (\GPT) are capable of logical reasoning with distorted facts, which we call \longdupe (\dupens).
\dupe presents a unique challenge to \GPT since they typically rely on their parameters, which encode mostly accurate information, to reason and make inferences. 
However, in \dupens, \GPT must reason over manipulated or falsified evidence present in their prompts, which can result in false conclusions that are valid only under the manipulated evidence.
Our goal with \dupe is to determine whether \GPT can arrive at these false conclusions and identify whether the dominant factor influencing the deduction process is the encoded data in the parameters or the manipulated evidence in the prompts.
To evaluate the \dupe capabilities of \GPTns, we create a \DUPED version of the StrategyQA dataset, where facts are manipulated to reverse the answer to the question.
Our findings show that even the most advanced GPT models struggle to reason on manipulated facts -- showcasing poor \dupe skills -- with accuracy dropping by 45$\%$ compared to the original dataset.
We also investigate prompt settings inspired from student simulation models, which mitigate the accuracy drop to some extent.
Our findings have practical implications for understanding the performance of \GPT in real-world applications such as student simulation models that  involve reasoning over inaccurate information.


\end{abstract}

\section{Introduction}
\begin{figure*}[ht]
    \centering
    \includegraphics[scale=0.41]{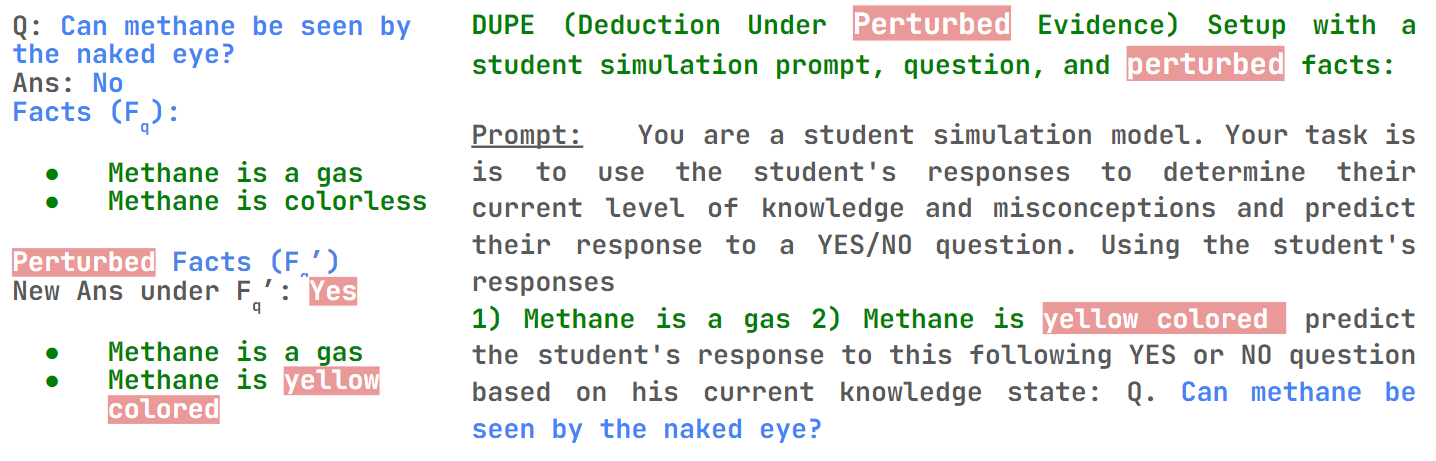}
    \caption{Setup of the \longdupe (\dupens)  reasoning framework. 
    On the left is a question-fact pair in StrategyQA dataset. To test \dupe skills of a model, we change facts provided with each question such that the  response to the question flips.
    On the right is a prompting setup to probe \dupe skills of \GPTns. We use a custom prompt tailored to student simulation setting that takes in the input question, perturbed (\dupedns) facts, and requests a \textit{yes/no} response from \GPTns.
    Perturbed facts represent a realistic student simulation setting since they mirror the inaccurate nature/ misconceptions of students' responses.
    }
    \label{fig:ckt}
\end{figure*}

Over the last several years, Transformer models have played a significant role in shaping the field of Natural Language Processing (NLP) \citep{attention,bert,roberta,gpt3,gpt3human,openai2023gpt4}.
Their exceptional ability to reason across a broad range of NLP tasks \cite{reasoningllm1,reasoningwpromptingllm2,gpt4sparkai} has been a key factor contributing to their success.
The success of \GPT on challenging datasets like HellaSwag \cite{zellers2019hellaswag}, AI2 Reasoning
Challenge (ARC) \cite{clark2018think}, WinoGrande \cite{sakaguchi2021winogrande}, and GSM-8K \cite{gsm8k} is a testament to their advanced reasoning skills and their potential to address challenging NLP tasks.


In this paper, we investigate the reasoning abilities of \GPT models under a novel paradigm we dub \longdupe (\dupe for short). 
By testing \GPTns' capacity to reason with flawed or perturbed evidence, we aim to determine whether \GPT can generate logically sound yet erroneous conclusions when presented with misleading information.
Strong \dupe skills are critical in NLP applications like student simulations \cite{dkt,okt}, where models simulate student responses to understand how they may respond in certain scenarios.
As student responses often contain inaccuracies and misconceptions, it is important for a model to analyze and utilize these inaccuracies and misconceptions as evidence to arrive at the same conclusion as the student.
For instance, a student may have the misconception that the heavier an object is, the faster it falls, leading them to conclude that a bowling ball will fall faster than a ball bearing.
If we provide \GPT with evidence that a heavier object falls faster, would \GPT also arrive at the conclusion that a bowling ball will fall faster than a ball bearing?
We introduce \dupe as our approach to investigate this question.
 
\begin{figure*}[t]
    \centering
    \includegraphics[scale=0.38]{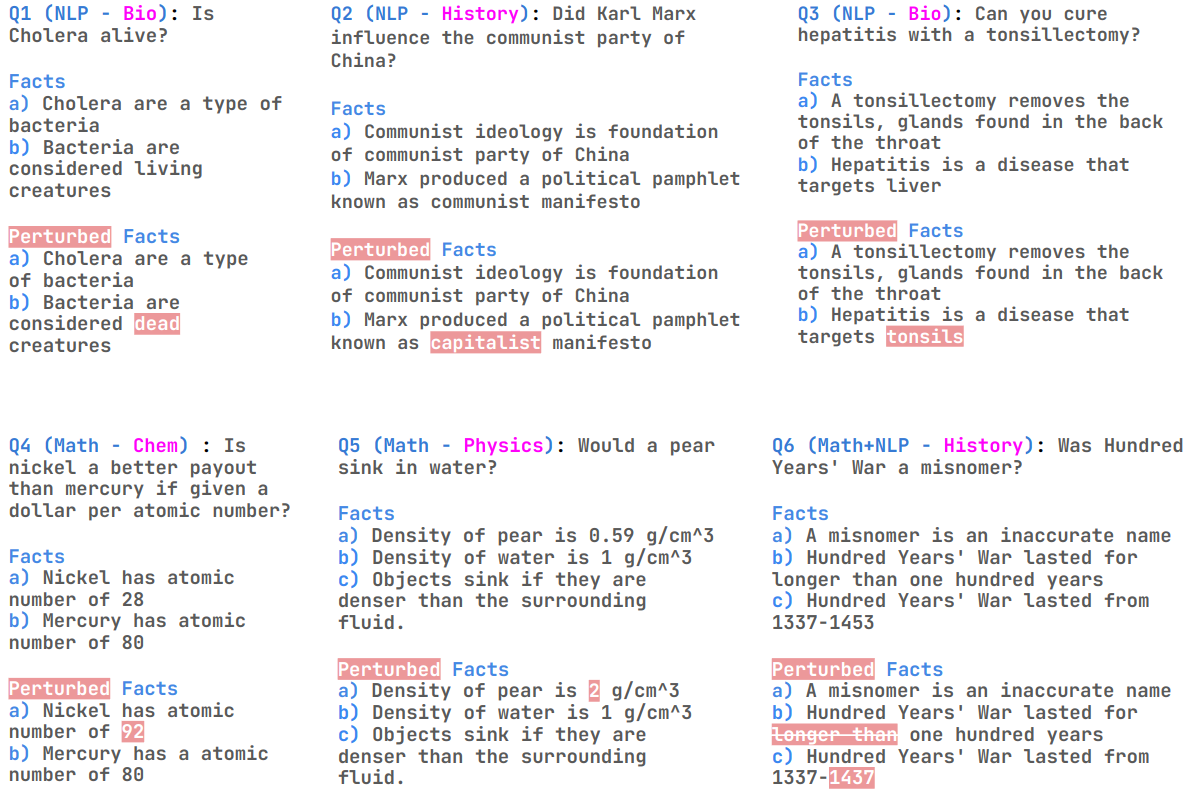}
    \caption{Six examples from our \DUPEDns-StrategyQA dataset. 
    We flip the answer to a \textit{yes-no} question by altering facts provided with each question.
    First three questions on the top are examples of natural language perturbations, while the bottom three questions involves manipulating numerical digits.
    The \duped version was designed with minimal modifications to the facts, usually involving only one to two word changes in the original facts.
    Additionally, we refrained from using explicit negation words like \textit{not}.
    }
    \label{fig:dupe_nlp}
\end{figure*}

\textbf{Contributions:} 
This paper develops a novel reasoning paradigm -- \longdupe (\dupens) -- to examine whether \GPT arrive at different conclusions when presented with distorted initial facts.
To test the \dupe capabilities of \GPTns, we create a \DUPED  version of StrategyQA dataset (Figures~\ref{fig:ckt}, \ref{fig:dupe_nlp}).
StrategyQA \cite{strategyQA} is an open-domain QA dataset that is characterized by its explicit provision of the necessary facts required to answer each \textit{yes-no} question.
In the \duped version of the dataset, we manipulate the facts provided in a way that results in a different answer to the original question.

Our findings reveal that state-of-the-art \GPTns,  , including GPT3.5 and GPT4, struggle significantly on the newly introduced \dupedns-StrategyQA dataset.
The accuracy of these models dropped drastically by approximately $45\%$, falling from an impressive $91.9\%$ on the original dataset to only $46.7\%$ on the \dupedns-StrategyQA dataset. 
In addition, we conduct an ablation study on the \dupedns-StrategyQA dataset by categorizing it into two distinct parts based on the type of manipulation used -- one involving language perturbations and the other involving mathematical manipulations.
Furthermore, our results demonstrate that the accuracy drop can be mitigated by using prompt settings inspired by student simulation models.
This approach reduced the accuracy drop to $29\%$, with the models achieving an accuracy of $62.7\%$ on the \dupedns-StrategyQA dataset.
Our findings carry crucial implications for practical \GPT applications, particularly in the realm of student simulation models that demand reasoning over erroneous information.


\section{Methodology, Dataset, and Prompting}
\label{sec:dataset}
\begin{table*}[t]
\centering
\resizebox{0.95\textwidth}{!}{
\begin{tabular}{|c|c|c|c|c|c|}
\hline
 \textbf{Dataset} & \textbf{Model} & \textbf{Prompt} & \textbf{Accuracy (Overall)} & \textbf{Accuracy (NLP)} & \textbf{Accuracy (Math)} \\ \hline
StrategyQA  & GPT3.5 & P1 & 84.6 & 94.1 & 74.4 \\ \hline
\DUPEDns-StrategyQA & GPT3.5 & P1 & 38.6 (46.0↓) & 35.4 (58.7↓) & 42.0 (32.4↓) \\ \hline
StrategyQA  & GPT4 & P1 & 91.9 & 94.1 & 89.4 \\ \hline
\DUPEDns-StrategyQA & GPT4 & P1 & 46.7 (45.2↓) & 43.8 (50.3↓) & 50.0 (39.4↓) \\ \hline
\DUPEDns-StrategyQA & GPT4 & P2 & 62.7 (29.2↓) & 63.1 (31.0↓) & 62.2 (27.2↓) \\ \hline
\end{tabular}
}
\caption{We evaluate the \dupe capabilities of the two largest  GPT models under two different prompt settings using the \dupedns-StrategyQA dataset.
Prompt P1 asks GPT models to answer a question based on provided evidence.
Under Prompt P1 setting, both GPT3.5 and GPT4 perform poorly on \duped version of the dataset with around $45\%$ accuracy drop.
We also find that both models are more robust to mathematical perturbation compared to natural language perturbations.
Prompt P2 is inspired from student simulation settings.
P2 primes the models that evidence provided may be incorrect.
We find that prompt P2 achieves better accuracy than Prompt P1 by $16.0$ points for GPT4, but we still see a substantial $29.2\%$ drop in accuracy compared to GPT4's accuracy on original dataset.
}
\label{tab:dupe}
\end{table*}

In this section, we overview the 
\dupe reasoning framework, provide details on the DUPEd version of AllenAI's StrategyQA dataset, and then explore customized prompt settings designed to assess the \dupe skills of \GPTns.

\subsection{\dupe} 
Given a \textit{true-false} question $q$, the correct response $r_q \in \{{\sl true,false}\}$ and facts $F_q$ that determine the truth or falsehood of $Q$ ($r_q$), we change $F_q$ to  $F_q'$ s.t.\ the correct response to $q$ flips to $\neg r_q$ under altered facts $F_q'$,
\begin{align}
\begin{split}
    \tr{DUPE}\big((q, F_q, r)\big) = (q, F_q', r') \\
    \tr{s.t.} \; r' = \neg r \;, \; {\rm edit}_{\rm dist}(F_q, F_q') < \tau,
\end{split}
\label{eq:dupeeq}
\end{align}
where ${\rm edit}_{\rm dist}$ ensures that the edit distance between the fact strings $F_q$ and  $F_q'$ is less than a threshold $\tau$.
The threshold $\tau$ is generally set to two to three words to ensure minimal changes to underlying facts (examples in figure~\ref{fig:dupe_nlp}).
The new \DUPEDns-tuple $(q, F_q', r')$ can be used to probe the \dupe capabilities of \GPT as shown in Figure~\ref{fig:ckt}.

\subsection{\dupedns-StrategyQA}
We use AllenAI's StrategyQA dataset \cite{strategyQA} to assess the \dupe skills of \GPTns.
StrategyQA dataset provides explicit facts for answering open-domain questions.
We create a \duped version of StrategyQA dataset composed of a total of 325 examples, of which 173 introduce natural language perturbations, while the remainder introduce mathematical errors (refer to examples in figure~\ref{fig:dupe_nlp}).

While designing the \duped version, we were careful to modify the facts in the most minimal way possible
As a result, we made a conscious effort to only alter one or two words in the original facts whenever possible, in order to preserve the overall meaning and context of the original question.
Additionally, we refrained from using explicit negation, such as the word \textit{not}, to modify the facts, since our intent is not to evaluate the reasoning proficiency of \GPT in handling negation.

\subsection{Student Simulation and Prompt Design}
\label{sec:prompt}
\dupe is highly relevant to {\em student simulation models} \cite{dkt,sonkar2020qdkt,okt}, which are widely used in education and cognitive psychology research.
These models help in predicting and understanding student responses to various tasks, and thus their ability to reason over false information is critical to their success.
Given this strong connection between simulation models and \dupens, these models can inspire innovative approaches to prompt design, which can be used to probe \dupe skills of \GPT \cite{reasoningwpromptingllm2,promptgpt1}.
An example of such a prompt is illustrated in figure~\ref{fig:ckt} and section~\ref{sec:setup}.


\textbf{\dupe and Counterfactual Reasoning:}
Counterfactual reasoning and student simulation models require different types of reasoning. 
In counterfactual reasoning, the focus is on exploring hypothetical scenarios that may or may not correspond to actual reality. 
The fact that the information being considered is hypothetical or counterfactual is usually known beforehand.

In contrast, a student simulation model needs to reason about both true and false information, and may not know beforehand whether the information being considered is true or false.
For example, in figure~\ref{fig:dupe_nlp}, the model lacks prior knowledge about which facts are true and which ones are perturbed.
The model must identify incorrect answers from the student to make inferences about future questions, which requires robust and nuanced reasoning capabilities beyond those needed for counterfactual reasoning.


\section{Experiments}
\label{sec:exp}
\label{sec:setup}
We evaluate the \dupe capabilities of the two largest GPT models -- GPT3.5 (version gpt-3.5-turbo-0301) and the latest GPT4 model (version gpt-4-0314) -- via  experiments under two different prompt settings, P1) ``You are a question answering model. Your task is reason on provided evidence to answer a YES or NO question'', and P2) ``You are a student simulation model. Your task is reason on student's responses to accurately measure the student's current knowledge state and predict the student's response to a YES or NO question based on the student's current knowledge state'' from section \ref{sec:prompt}.
An example is illustrated in Figure~\ref{fig:ckt}.

\subsection{Main Results}
In the prompt setting P1, both GPT3.5 and GPT4 performed poorly on the \duped version of the dataset, with a decrease in accuracy by $46.0\%$. and $45.2\%$ respectively.
As expected, the latest GPT4 model demonstrates superior performance to GPT3.5 on both the original and the \duped StrategyQA dataset.

\subsubsection{Student Simulation Prompt}

Prompt P2 inspired by student simulation setting informs/ primes the models that the provided evidence may be incorrect since the evidence reflects the erroneous nature of students' responses.
We found that prompt setting P2 performs significantly better than P1 by a margin of $16.0\%$ for the GPT4 model.
However, there was still a significant $29.2\%$ drop in accuracy compared to GPT4's performance on the original dataset.

\subsubsection{Language vs.\ Math Perturbations}
While curating the \dupedns-StrategyQA dataset, we divided the perturbations introduced into two distinct categories - one that involved language perturbations, while the other manipulated mathematical information (see figure~\ref{fig:dupe_nlp}).
Our finding suggest that both GPT models are more resilient to math perturbations compared to language perturbations.
E.g. for GPT3.5 there was accuracy drop of $58.7\%$ and $32.4$ for language and math Perturbations respectively, while for GPT4 the  accuracy drops were $50.3\%$ and $39.4$.

\subsection{Root Cause of Poor \dupe Skills}
To explain the GPT models' poor performance on the \DUPED dataset, we need to identify the main factor influencing their reasoning process, i.e., whether it is the encoded information in parameters or the manipulated evidence in prompts. 
Recent studies have shed light on this issue, suggesting that factual information encoded in the parameters of \GPT plays a dominant role in governing the generated output.
For instance, the feed-forward layers in transformer models function as key-value memories, which implies that they encode factual information, as noted by \citet{ffnkvmemories}. 
Moreover, \citet{meng2022locating} demonstrated that localized computations, such as Rank-One Model Editing (ROME), can modify these factual associations, leading to alternative conclusions. 
These findings suggest that the encoded information in parameters has a significant impact on \GPTns' reasoning process; further investigation is left for future work.

\section{Conclusions}
In this paper, we have introduced a new reasoning paradigm we call  \longdupe (\dupe for short).
Through \dupe, we have assessed the ability of \GPT models to arrive at logically sound yet erroneous conclusions when faced with distorted initial facts.
Our study, which used a carefully curated dataset to evaluate \dupe abilities, has revealed that even the most advanced GPT models struggle with logical reasoning in the presence of falsified information. 
Moving forward, we plan to investigate into the performance of different \GPT with our dataset in varied prompt settings.

\section{Limitations}
Due to limitations in both financial and computational resources, we had to limit our testing to only the most advanced \GPT -- the GPT models.
Consequently, we directed our attention towards developing a dataset for evaluating proposed reasoning scenarios.
As a result of these limitations, we chose to focus specifically on the evaluation of the two largest models offered by OpenAI.
While we recognize that other \GPT may produce different outcomes, we believe that our dataset 
could serve as a valuable resource for further research into the capabilities and limitations of \GPT.

\bibliography{anthology,custom}
\bibliographystyle{acl_natbib}



\end{document}